\documentclass[conference]{IEEEtran}
\IEEEoverridecommandlockouts
\usepackage{amsmath,amssymb,amsfonts}

\usepackage{multirow}
\usepackage{algorithm}
\usepackage{algpseudocode}
\usepackage{graphicx}
\usepackage{textcomp}
\usepackage{xcolor}

\def\BibTeX{{\rm B\kern-.05em{\sc i\kern-.025em b}\kern-.08em
    T\kern-.1667em\lower.7ex\hbox{E}\kern-.125emX}}
\begin{document}

\title{Multi-objective Optimisation of Multi-output Neural Trees}

\author{\IEEEauthorblockN{Varun Ojha}
\IEEEauthorblockA{\textit{University of Reading}\\
Reading, United Kingdom \\
0000-0002-9256-1192}
\and
\IEEEauthorblockN{Giuseppe Nicosia}
\IEEEauthorblockA{\textit{University of Cambridge}\\
Cambridge, United Kingdom \\
{0000-0002-0650-3157}} 
}

\maketitle

\begin{abstract}
We propose an algorithm and a new method to tackle the classification problems. We propose a \textit{multi-output neural tree} (MONT) algorithm\footnote{Source code: https://github.com/vojha-code/Multi-Output-Neural-Tree}, which is an evolutionary learning algorithm trained by the non-dominated sorting genetic algorithm (NSGA)-III. Since evolutionary learning is stochastic, a hypothesis found in the form of MONT is unique for each run of evolutionary learning, i.e., each hypothesis (tree) generated bears distinct properties compared to any other hypothesis both in topological space and parameter-space. This leads to a challenging optimisation problem where the aim is to minimise the tree-size and maximise the classification accuracy. Therefore, the Pareto-optimality concerns were met by hypervolume indicator analysis. We used nine benchmark classification learning problems to evaluate the performance of the MONT. As a result of our experiments, we obtained MONTs which are able to tackle the classification problems with high accuracy. The performance of MONT emerged better over a set of problems tackled in this study compared with a set of well-known classifiers: multilayer perceptron, reduced-error pruning tree, na\"{i}ve Bayes classifier, decision tree, and support vector machine. Moreover, the performances of three versions of MONT's training using genetic programming, NSGA-II, and NSGA-III suggests that the NSGA-III gives the best Pareto-optimal solution. 
\end{abstract}


\section{Introduction}

Learning from data is essentially a search process by which we search for a hypothesis (a trained model) from a hypothesis space that maps (fits) the given input data to its target  output as good as possible (the high accuracy on test data). A learning algorithm like multilayer perceptron's architecture and parameter tuning are the efforts to find a hypothesis that fits the data well~\cite{annBpRumelhart1986,ojha2017metaheuristic}. Similarly, there is a variety of hypothesis selection possible. Such examples are decision tree~\cite{quinlan2014c4}, reduced error pruning tree~\cite{kohavi2002data}, na\"{i}ve Bayes classifier~\cite{john1995estimating}, and support vector machine~\cite{chang2011libsvm}.

In this study, our effort is to take advantages of the evolutionary processes for designing a new method for searching a hypothesis (an evolutionary learning algorithm) that fits well on a variety of datasets. Therefore, we propose a multi-output neural tree (MONT) algorithm, which resembles a tree data structure whose nodes are neural nodes similar to the nodes of a multilayer perceptron. The tree exploits the genetic programming (GP)~\cite{schmidt2009solving} and non-dominated sorting genetic algorithm (NSGA) frameworks II~\cite{deb2000fast} and III~\cite{deb2013evolutionary} to evolve from data, at independent instances.

The proposed algorithm MONT is an innovation from the early tree-based learning algorithms such as a flexible neural tree where a tree-like-structure was optimised by using probabilistic incremental program evolution~\cite{chen2005time} and heterogeneous flexible neural tree (HFNT)~\cite{ojha2017ensemble} where a tree-like-structure was optimised by NSGA-II. Similar to these two approaches, in~\cite{ojha2017multiobjective}, a fuzzy inference system enabled hierarchical tree-based predictors was illustrated. In~\cite{bouaziz2013hybrid}, a tree-based algorithm was evaluated on beta-basis function as a neural node. Among these algorithms, the proposed MONT algorithm closely linked to HFNT. Hence, a comparison of MONT with HFNT is presented in this research. 

These early versions of the tree-based algorithms are limited to binary class classification since the root node reports the output. Hence, these algorithms worked in a multi-input-single-output fashion. For the multi-class classification, these algorithms need to be repeated for each class separately, which results in as many as trees as the number of classes. 

Our proposed algorithm eliminates this limitation by using a single tree to learn for multiple classes. In MONT algorithm, each child of the tree's root-node is formulated as the class output. Hence, the proposed MONT works in multi-input-multi-output fashion and treats binary classification as two-class classification. The results show that the competitive nature of the evolutionary process improve performance. Moreover, our proposed method applies NSGA-III combined with hypervolume inductor analysis~\cite{fonseca2006improved} to obtain the best neural trees serving Pareto-optimality for an evolutionary learning process. The contributions of this study are as follows: %
\begin{itemize}
    \item A new algorithm called MONT is designed for classification tasks, specifically aims at adapting multi-class.
    \item A new method is proposed for neural trees generations.
    \item A Pareto-optimality of evolutionary learning processes was investigated using hypervolume indicator analysis. 
    \item A comprehensive analysis of the MONT's (trained with NSGA-III) performance compared with other algorithms and with MONT's other two training version GP, NSGA-II is presented.
\end{itemize}
The rest of the paper is organised as follows: Section~\ref{sec:mont} describes the multi-class classification problem and the basic architecture, principles, and properties of MONT algorithm. Section~\ref{sec:mo_ec} describes the evolutionary learning processes and framework for constructing optimal neural trees. Section~\ref{sec:exp_setup} describes the experimental setup designed for the induction of a varied range of hypothesis, including MONT and well-known five algorithms over a diverse range of datasets. The results of the experiments are summarised in Section~\ref{sec:res} and discussed in Section~\ref{sec:dis}, followed by conclusions in Section~\ref{sec:con}.
\section{Multi-Objective-Multi-Output Neural Trees}
\label{sec:mont}
\subsection{Problem Statement} 
\label{sec:problem_stmt}
Let $ \mathcal{X} \in \mathbb{R}^d$ be an instance-space and let $ \mathcal{Y} = \{c_1, \ldots,c_r \} $ be a set of $ r $ labels such that label $ y \in \mathcal{Y} $ can be assigned to an instance $ \textbf{x} \in \mathcal{X} $. Hence, for a training set of instance-label pairs $ \mathcal{S} = \left(\textbf{x}_i, y_i \right)_{i=1}^{N} $, we face a multi-class learning problem. And, a hypothesis $ h $ from a set of hypothesis class $ H $ is induced that aims at reducing a cost function $ f(\cdot)$. A typical cost function is the classification error-rate:
\begin{equation}
\label{eq:error_rate}
f =  \frac{1}{N}\sum\limits_{i=1}^{N} (h(\textbf{x}_i) \ne y_i)
\end{equation}
where $ h(\textbf{x}_i)$ is the predicted output for an input instance $\textbf{x}_i = \left\langle x^i_1, x^i_2, \ldots, x^i_d \right\rangle$ and $ y_i \in \{c_1, \ldots,c_r \}$ being its target class.
\subsection{Multi-Output Neural Trees}
\label{sec:nt}
Multi-output neural tree (MONT) takes a tree-like structure where the root node takes as many as child nodes as the number of classes it is to be induced on. Each child node of the tree's root predicts a class, and each of them is a subtree of the MONT classifier.

\begin{figure}	
    \centerline{\includegraphics[scale = 0.55]{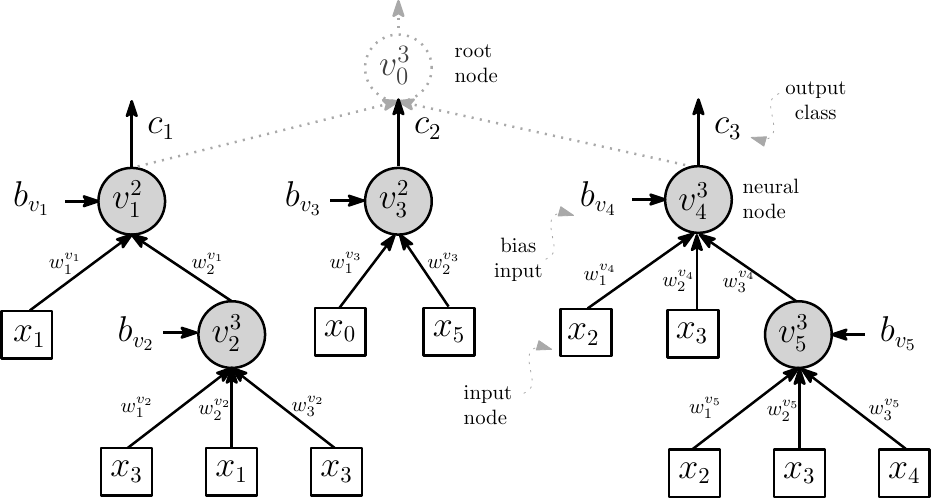}}
    \caption{Representation of a multi-output neural tree for a three-class problem with classes $ c_1, c_2, $ and $ c_3 $. The immediate child nodes $ v_1, v_3, $ and $ v_4 $ of the root node $ v_0 $ are the output class nodes. The other internal nodes or leaf nodes construct subtrees for each root child (the respective output class). This tree takes its input from the set \{$x_1, x_2, \ldots, x_5$\}. The link $ w_i^{v_j} $ between nodes are neural weights.}
    \label{fig:mont}
\end{figure}

Mathematically, MONT, $ G $, is an $ m $--ary rooted tree with one node designated as the root node and each node takes at least $ m \ge 2$ child nodes except for the leaf node that has no child node. Hence, for a tree depth $ p $, MONT takes a maximum of $ (2^{p+1}-1) \le n \le (m^{p+1}-1)/(m-1)$ nodes (including the number of internal nodes $ K = |V| $ and the leaf node $ L = |T|$). A MONT, $ G $, is denoted as:
\begin{equation}
\label{eq_mont}
G = V \cup T = \left\lbrace v^j_1,v^j_2,\ldots,v^j_K \right\rbrace  \cup \left\lbrace t_1, t_2,\ldots, t_L \right\rbrace 
\end{equation}
where $ k $-th node $ v^j_k \in V$ is an internal node and receives $ 2 \le j \le m $ inputs from its child nodes. The $ k $-th leaf node $ t_k \in T$ has no child and it contains an input $ x_i \in  \{x_1,x_2,\ldots,x_d\}$. An example of a MONT is shown in Fig.~\ref{fig:mont}.  

Fig.~\ref{fig:mont} is an example of three class problem, where the root node $ v_0^3 $ takes three child nodes $ v_1^2 $, $ v_2^2 $, and $ v_3^3 $ representing three classes $ c_1 $, $ c_2 $, and $ c_3 $. Each child node of the root is a full $ m $-ary subtree. For example, node $ v_1^2 $ takes two child nodes and the node $ v_1^3 $  takes three child nodes. The number of child nodes and the size of subtrees is governed by an evolutionary learning process. The leaf nodes of the MONT are the input nodes that takes an input feature $ x \in \textbf{x}$.

\begin{figure}	
    \centerline{\includegraphics[scale = 0.6]{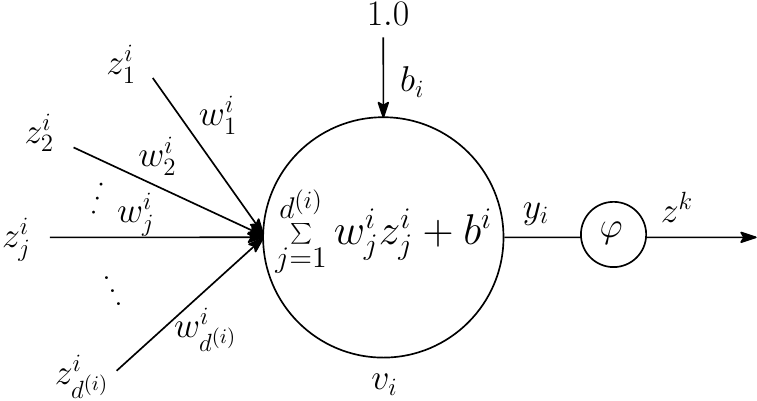}}
    \caption{ Illustration of a computational (neural) node. The variable $ d_{v_i} $ indicates the number of inputs $ z^i_j $ and weights $ w^i_j $ received at the $ i $-th node $ v_i $, the variable $ b_{i} $ is the bias at the $ i $-th node and the variable $ z^k $ is the output of the $ i $-th node squashed by an activation function $ \varphi(y_i) $.}
    \label{fig:sing_node}
\end{figure}

The internal nodes of the MONT are neural nodes. Each internal node has an activation function (e.g., Gaussian, sigmoid, tangent hyperbolic) and behave similarly to a node in multilayer perceptron. Fig.~\ref{fig:sing_node} is an example of the $ i $-th MONT's neural node that receives the inputs from its child nodes (higher tree-depth) and produces an output for its parent node (e.g., $ k $-th node in lower tree-depth).

A MONT resembles an expression tree, and the computation of MONT takes place as per depth-first-search in pre-order fashion (Fig.~\ref{fig:mont}). Hence, the \textit{time complexity} of MONT for computing its output is $ \mathcal{O}(n) $, where $ n $ is the number of nodes in the tree.

\subsection{Multi-Objective Optimisation of Neural Trees}
\label{sec:mo_ec}
In our designed evolutionary process, an evolutionary algorithm like NSGA-III searches through all possible combination of MONTs, which is roughly close to Catalan number $ C_n = 1/[(m-1)(n+1)].{\binom{m.n}{n}}$, and for MONT, $ n $ is $ \ge 7 $ and only 1 tree-structure (shape) possible for $ n = 7$ since MONT takes at least two classes and each class takes at least two inputs. Hence, 1 root node, 2 child nodes and 4 leaf node, total 7 nodes can only be arranged in one unique structure as per MONT's definition (Section~\ref{sec:nt}). However, the parameters of the tree's edges and the node's function further increase the search-space size. 

Felsenstein presented a theory for all possible combinations of tree-structure for an $ m $--ary tree that has a total $ L = m^p $ labelled leaf nodes~\cite{felsenstein1978number}, and, for $ L $ labelled leaf nodes, a total number of possible tree-structures arrangements (combinations) shown for an evolutionary process is $ (2L - 3)!/(2^{L-2}(L-2))!$~\cite{felsenstein1978number}. Therefore, an evolutionary processes search through such a vast hypothesis search-space to obtain an optimal tree.

In our method, we used three training versions for MONT: NSGA-III~
\cite{deb2013evolutionary}, NSGA-II~\cite{deb2000fast}, and GP~\cite{schmidt2009solving}. MONT takes genetic operators such as crossover (of subtrees)  and mutation defined in~\cite{ojha2017ensemble}. In~\cite{ojha2017ensemble}, the following forms of mutation is defined for the tree's mutation: 1) deletion of a randomly selected leaf node, 2) replacement a randomly selected leaf node, 3) replacement of a randomly selected function node by a leaf node or a new subtree. Using the mentioned genetic operators, the MONT follows the typical evolutionary computation steps as per Algorithm~\ref{algo:mont_ec} for its training~\cite{goldberg1989genetic}. For MONT's multi-objective training, we supplied two objectives to be minimised: classification  error-rate $ f_1 $ and tree-size $ f_2 $. The error-rate $ f_1 $ is expressed as per~\eqref{eq:error_rate}. The objective tree-size $ f_2$ is $|G|$, i.e.,  the total number of nodes in a tree $ G $.
\begin{algorithm}
    \caption{Evolutionary Learning of MONT}
    \label{algo:mont_ec}
    \begin{algorithmic}[1]
        \Require{Initial population $ P_0 $ of randomly generated neural trees, objectives $ \mathcal{F} = [f_1,f_2]$ , data $ \mathcal{S} $, maximum evolutionary generations (termination criteria) $ g_{max} $.}
        \Ensure{Final population $ P_{g_{max}} $ of Pareto-optimal trees}
        \Statex
        \Function{Tree Evolution}{$P_0$, $\mathcal{F}, \mathcal{S}, g_{max}$.}
        \While{number of generation $ g $ reached $g_{max}$}
        \State selection: parent trees for crossover and mutation
        \State generation: a new population $ Q $
        \State combined population: $ R = P_g + Q $
        \State evaluation: NSGA-II/III non-dominated sorting($ R $)
        \State survive: elitism/niching ($ P_{g+1}, size(P_0), R $)
        \EndWhile
        \State \Return { $ P_{g_{max}} $}
        \EndFunction
    \end{algorithmic}
\end{algorithm}

For MONT training, the algorithms GP, NSGA-II, and NSGA-III follow Algorithm~\ref{algo:mont_ec} with some differences in a few steps. Especially, they differ in line numbers 6 and 7 of Algorithm~\ref{algo:mont_ec}. In GP, no non-dominated sorting is performed. Instead, the recombined population $ R $ in line no. 5 is sorted according to single objective $ f_1 $ (line no. 6 of Algorithm~\ref{algo:mont_ec}). Whereas, NSGA-II and NSGA-III both follows {non-dominated sorting} as per~\cite{deb2000fast}. However, NSGA-II and NSGA-III differ in line no. 7 of Algorithm~\ref{algo:mont_ec}.  NSGA-II performs \textit{elitism} based on \textit{crowding distance} of the individuals computed based on each objective~\cite{deb2000fast}. On the other hand, NSGA-III performs \textit{niching}. The niching operation takes advantage of a predefined set of reference points placed on a normalised hyperplane of an $ M $-dimensional objective-space~\cite{das1998normal}, where each individual in the population is associated to a reference point~\cite{deb2013evolutionary}. Moreover, the total number of reference points depends on the division of each objective axis. Both, the crowding distance and niching in NSGA-II and NSGA-III respectively aims at preserving diversity in the population.

\subsection{Hypervolume Analysis for Pareto-Optimality}
\label{sec:hv_anl}
An evolutionary process (e.g. NSGA-II or NSGA-III) for two objectives gives a non-dominated set of solutions. A non-dominated solution is the one for which no one objective function can be improved without a simultaneous detriment to at least one of the other objectives~\cite{deb2000fast}. The non-dominated solution is also known as the Pareto-optimal solution. Moreover, a set of such solution creates a Pareto-optimal front. 

In this study, we choose to compute the hypervolume indicator $ H_i $, which measures the dominance of Paret-front solutions on a geometric space (area for a 2D objective space) framed by the $ M $-dimensional objective-space with respect of a positive semi-axle. 
Hence, $ H_i $ measures the quality Pareto-optimal solutions set~\cite{fonseca2006improved}, and it is an indicator of the quality of the solutions obtained by two algorithms with respect to the same reference frame. 
We want hypervolume indicator index $H_i$ to be maximised. A greater value indicates that the overall performance of the algorithm is better with respect to another algorithm associated with a smaller hypervolume value. Moreover, the greatest contributing point in a hypervolume indicator analysis is the point that covers the largest area and that can be considered as the best solution~\cite{zitzler2003performance}.

\subsection{Experiment Set-Up}
\label{sec:exp_setup}
We designed our experiments to evaluate the performance of MONT algorithm on a set of different datasets each pertaining to distinct feature-space and a varied number of classes. The datasets used for the experiments were retrieved from the UCI machine learning repository\cite{lichman2013uci}. The details of the dataset are described in Table~\ref{tab:data}. These datasets were chosen because of their diversity in feature-space. For example, the dataset Australia and Heat have a mix of nominal and categorical attributes (features), the dataset Iris and Glass have real-values attributes, and the dataset Ionosphere has it every attribute within the range of -1.0 to 1.0. These make a single algorithm to perform equally well on each dataset difficult~\cite{wolpert1996lack}.
\begin{table} [b]
    \begin{center}   
        \renewcommand{\arraystretch}{1.2}
        {\caption{Descriptions of the Datasets Used in the Experiments.}
            \label{tab:data}}
        \begin{tabular}{llrrr}
            \hline
            Index & Name & Features & Samples & Classes\\
            \hline
            aus & Australia  & 14 &  691 & 2 \\
            hrt & Heart      & 13 &  270 & 2 \\
            ion & Ionosphere  & 33 &  351 & 2 \\
            pma & Pima       &  8 &  768 & 2 \\
            wis & Wisconsin  & 30 &  569 & 2 \\
            irs & Iris       &  4 &  150 & 3 \\
            win & Wine       & 13 &  178 & 3 \\
            vhl & Vehicle    & 18 &  846 & 4 \\
            gls & Glass      &  9 &  214 & 7 \\
            \hline
        \end{tabular}
    \end{center}
\end{table}

The performance of the proposed MONT (NSGA-III version) trained over these datasets was compared with the performance of five well-known algorithms bearing a differing characteristic and a tree-based algorithm heterogeneous flexible neural tree (HFNT)~\cite{ojha2017ensemble}. We used decision tree (DT)~\cite{quinlan2014c4}; multilayer perception (MLP)~\cite{annBpRumelhart1986}; reduced error pruning tree (REP-T)~\cite{kohavi2002data}; na\"{i}ve Bayes classifier (NBC)~\cite{john1995estimating}; and support vector machine (SVM)~\cite{chang2011libsvm}. We chose the state-of-the-art implementations of these algorithms from the Weka tool~\cite{hall2009weka}. 

Each of these chosen algorithms differs in their nature. Therefore, we expect them to perform differently over a diverse range of the dataset. Hence, we look for the average performance of MONT over a set of datasets in comparisons to the average performance of the chosen algorithms, as well as; we look for the comparisons between three different versions of neural tree (MONT) training processes and settings. These versions are: single objective GP based training, MONT$_1$, NSGA-II enabled multi-objective optimisation, MONT$_2$, and NSGA-III enabled multi-objective optimisation, MONT$_3$. Additionally, each of these training versions MONT$_1$, MONT$_2$, and MONT$_3$ also run for three different activation nodes: Gaussian, sigmoid, and tangent hyperbolic (tanh) functions. Hence, six training versions of MONT were evaluated.  

The three training versions MONT$_1$, MONT$_2$, and MONT$_3$ respectively assume GP, NSGA-II, and NSGA-III and follows the evolutionary process outlined in Algorithm~\ref{algo:mont_ec}. For the comparison of solutions obtained by these three versions of MONT training, a reference point 1.0 and 100 respectively indicating the worst test error-rate and worst tree-size was chosen for computing hypervolume indicator index $ H_i $.     

For the performance comparisons, we set-up a hold-out method of validation where datasets were randomly partitioned into 80\% training and 20\% test sets. For each dataset, 30 runs of algorithm training and testing results were collected. The parameter settings of the chosen algorithms were set to their default settings prescribed in Weka tool. A summary of the training parameters used for each algorithm are as follows:
\begin{itemize}
    \item MONT: iterations - 100; population - 50; max child nodes - 5; max tree height - 10; optimisers - GP, NSGA-II, and NSGA-III; crossover probability 0.5; mutation probability 0.5; NSGA-III's reference point division - 10.
    \item HFNT: iterations - 100; population - 50; max child nodes - 5; max tree height - 10; optimisers - NSGA-II; crossover probability 0.5; mutation probability 0.5.
    \item MLP: iterations - 500; number of hidden layer nodes - (features + classes)/2; learning rate - 0.3; momentum rate - 0.2; optimiser-  backpropagation.
    \item REP-T:  max tree height - unlimited; tree punning - no pruning; max feature at a node - 2.
    \item NBC: distribution function - normal distribution.
    \item DT: attributes information quality measure -  Gini index; tree punning - no pruning.
    \item SVM: kernel type at the nodes - redial basis function.
\end{itemize}

\section{Results}
\label{sec:res}
\subsection{Hypervolume Indicator Analysis and Tree Selection}
\label{sec:res_nt_hv}
\begin{figure}
    \centerline{\includegraphics[width=0.45\textwidth]{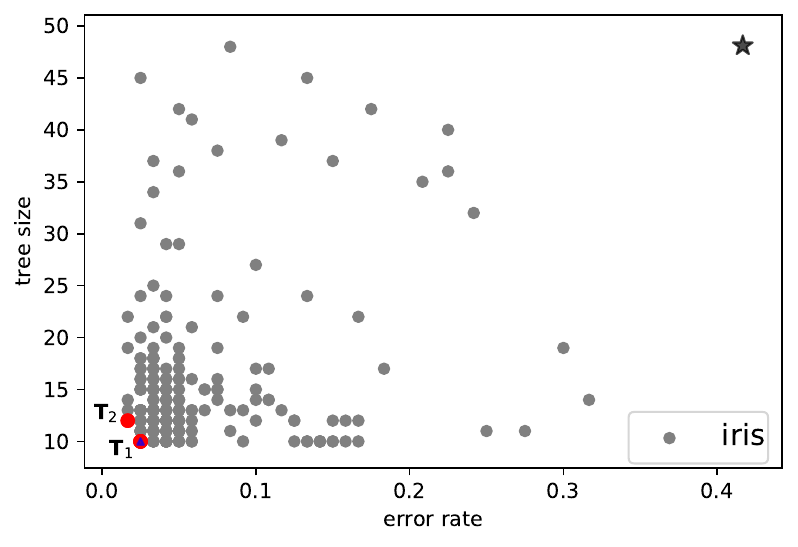}}
    
    \centerline{\includegraphics[width=0.45\textwidth]{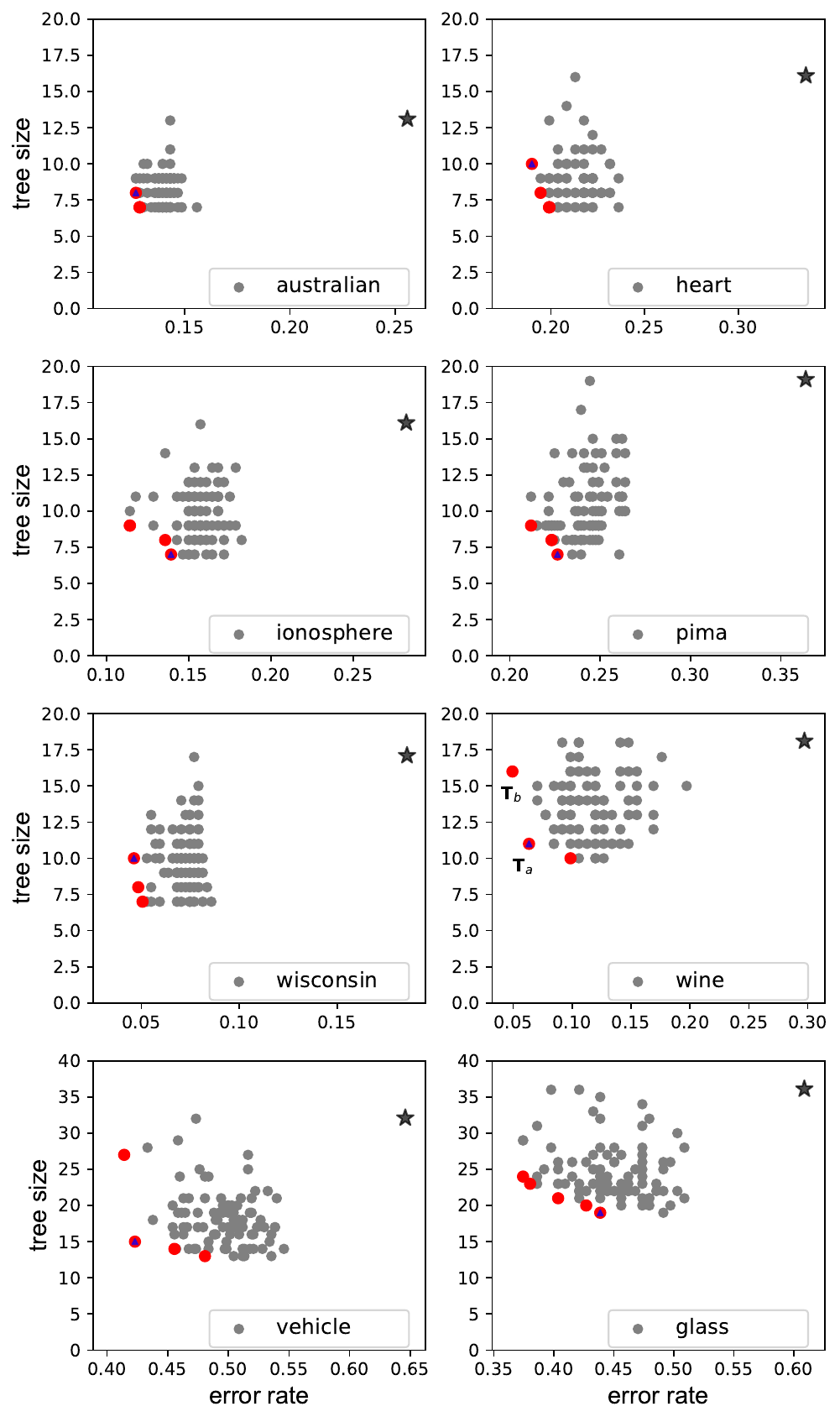}}

    \caption{Hypervolume analysis of MONT$_3$ over datasets (Iris top) and on aus, hrt, ion, pim, wis, win, vhl, and gls, respectively from top left to bottom right. The reference point marked in ``*'', blue (upside) triangle is the greatest contributing point. Pareto-front solutions are indicated by the red dots on the left bottom corner and other feasible solutions are indicated in gray dots. Select trees are marked T$_i$ (see iris and wine plots) and are shown in Fig~\ref{fig:nt_best_wine}.} 
    \label{fig:hv_plots}
\end{figure}

Each solution in the MONT's population has an error-rate and a tree-size associated with it which are conflicting objectives. For MONT's population, the Pareto-optimal solutions set can be evaluated by hypervolume indicator index $ H_i $ as discussed in Section~\ref{sec:hv_anl}. The hypervolume indicator analysis allow us to select the best solution from the MONT$_3$'s population for each dataset. Fig.~\ref{fig:hv_plots} shows hypervolume indicator analysis of the dataset over two objectives: neural tree-size against training error-rate. In Fig.~\ref{fig:hv_plots}, the blue (upside) triangle indicates the greatest contributing point with respect to the reference point (maximum tree-size and maximum error-rate of the population with an offset 0.1) indicated with symbol ``*.'' The Pareto-optimal solutions are indicated in red dots and other feasible solution are in gray.

The best neural tree (MONT$_3$) for each dataset was obtained using the hypervolume indicated analysis. The greatest contributing point was considered as the best MONT satisfying both objectives: tree-size and error-rate. The average test error-rates of the best trees of the 30 runs are compared with other algorithms (Section~\ref{sec:res_nt_vs_all}).

\begin{figure}
    \centering{
        \includegraphics[scale=0.3]{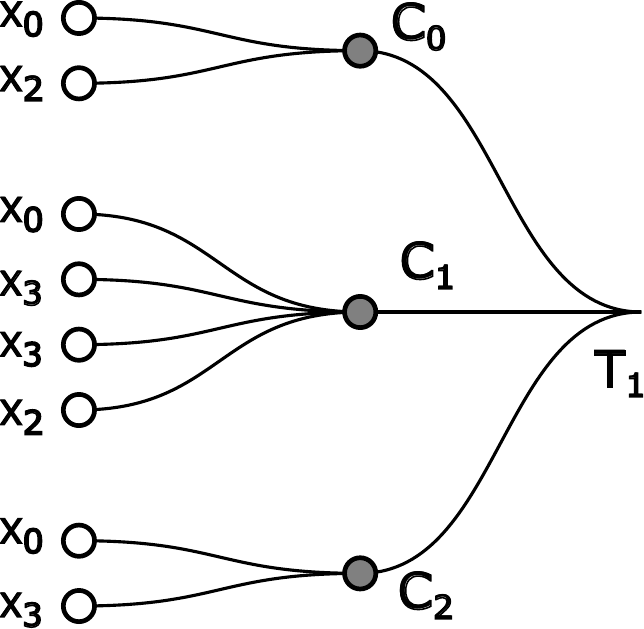}\quad\quad
        \includegraphics[scale=0.3]{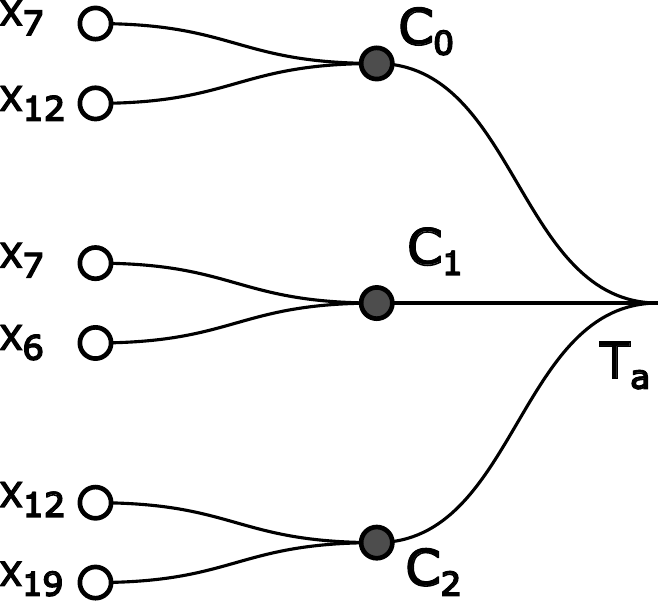}
        
        ~\\
        
        \includegraphics[scale=0.26]{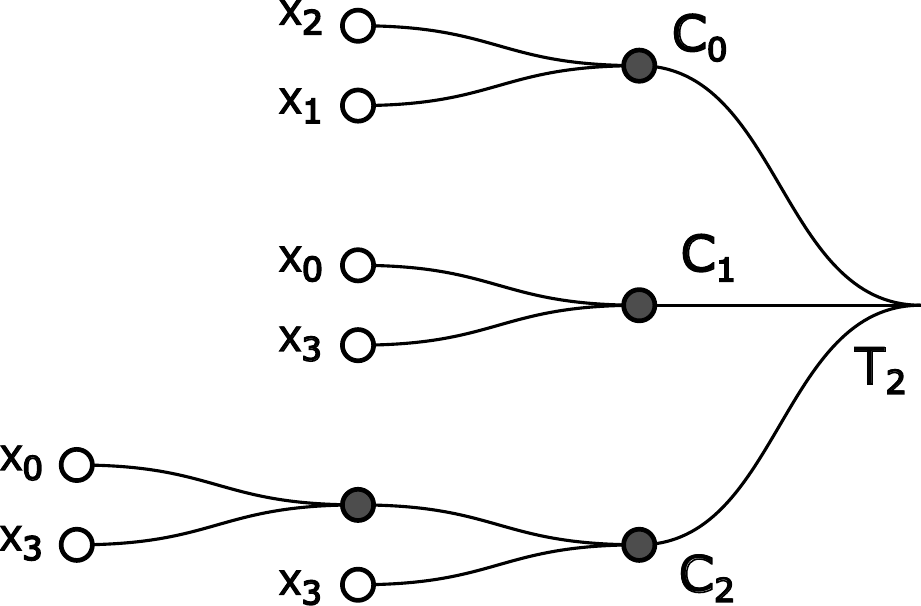}
        \includegraphics[scale=0.26]{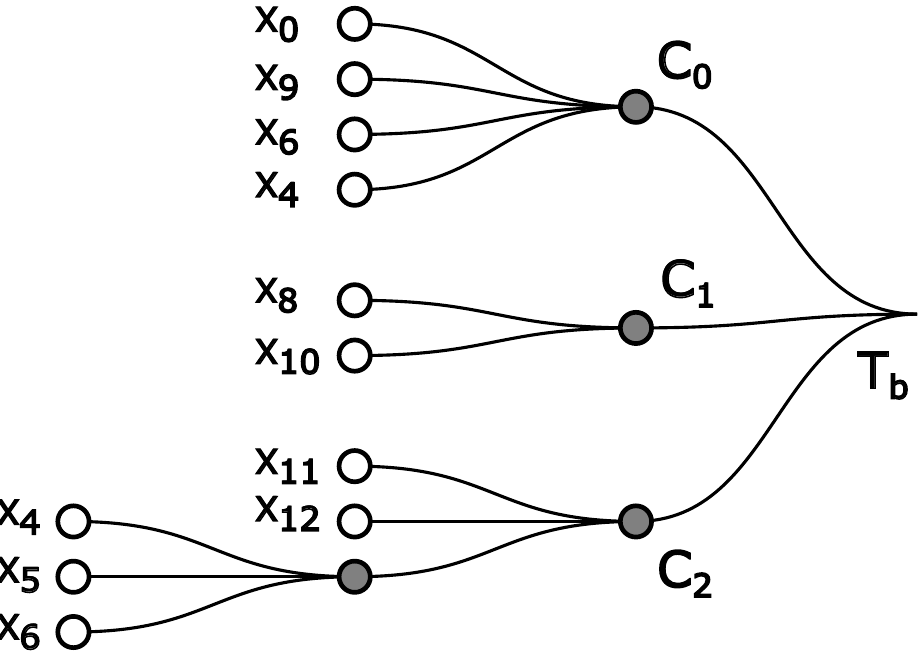}
    }
    \caption{Best performing trees (as per Fig.~\ref{fig:hv_plots}) for datasets Iris (T$_1$ and T$_2$ ) and Wine (T$_a$ and T$_b$). The shaded nodes are function nodes. Trees T$_1$, T$2$, T$_a$, and T$_b$ of respective datasets gives test error-rate 0.00, 0.013, 0.00, 0.167. } 
    \label{fig:nt_best_wine}
\end{figure}

\begin{figure}
    \centering{
        \includegraphics[width=0.4\textwidth]{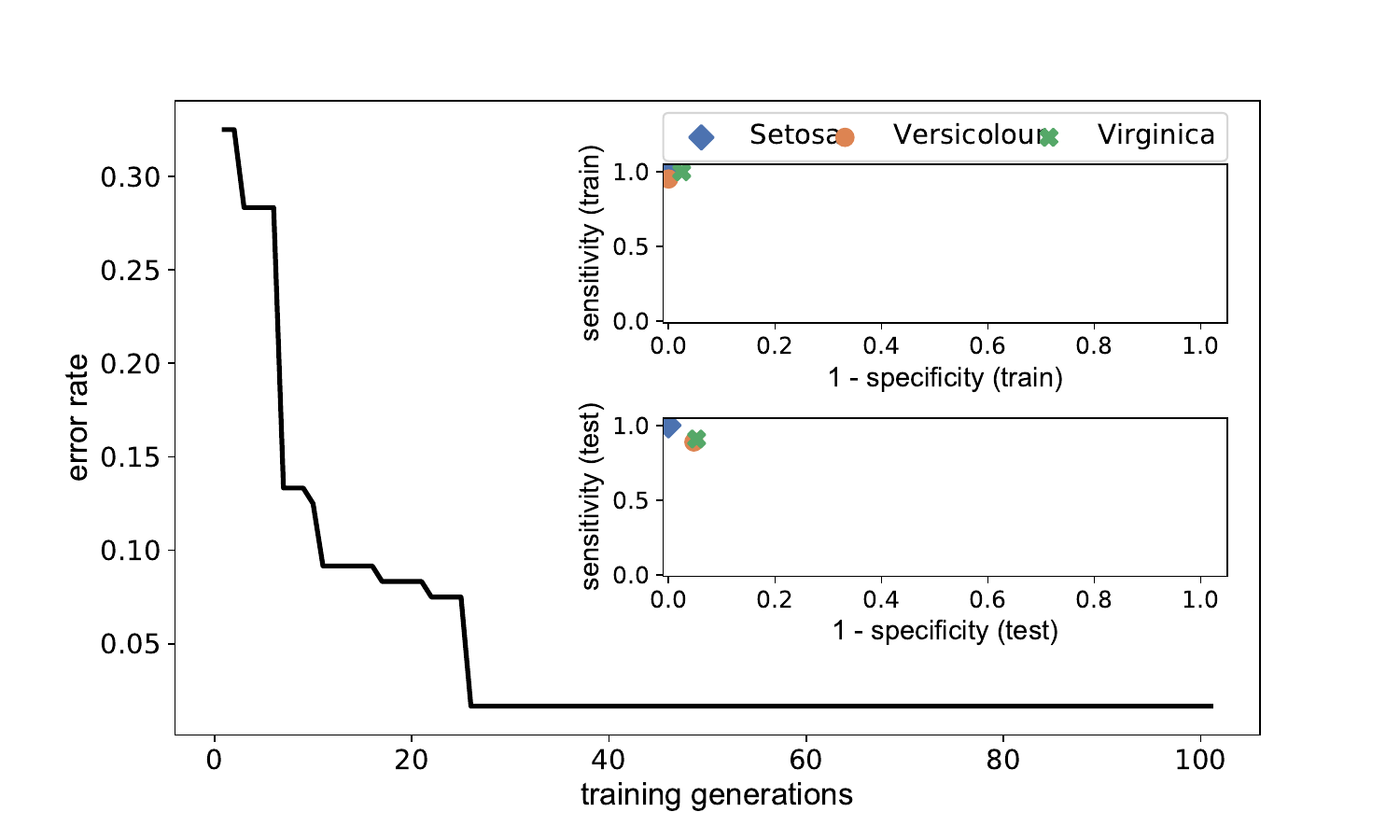}
        \includegraphics[width=0.4\textwidth]{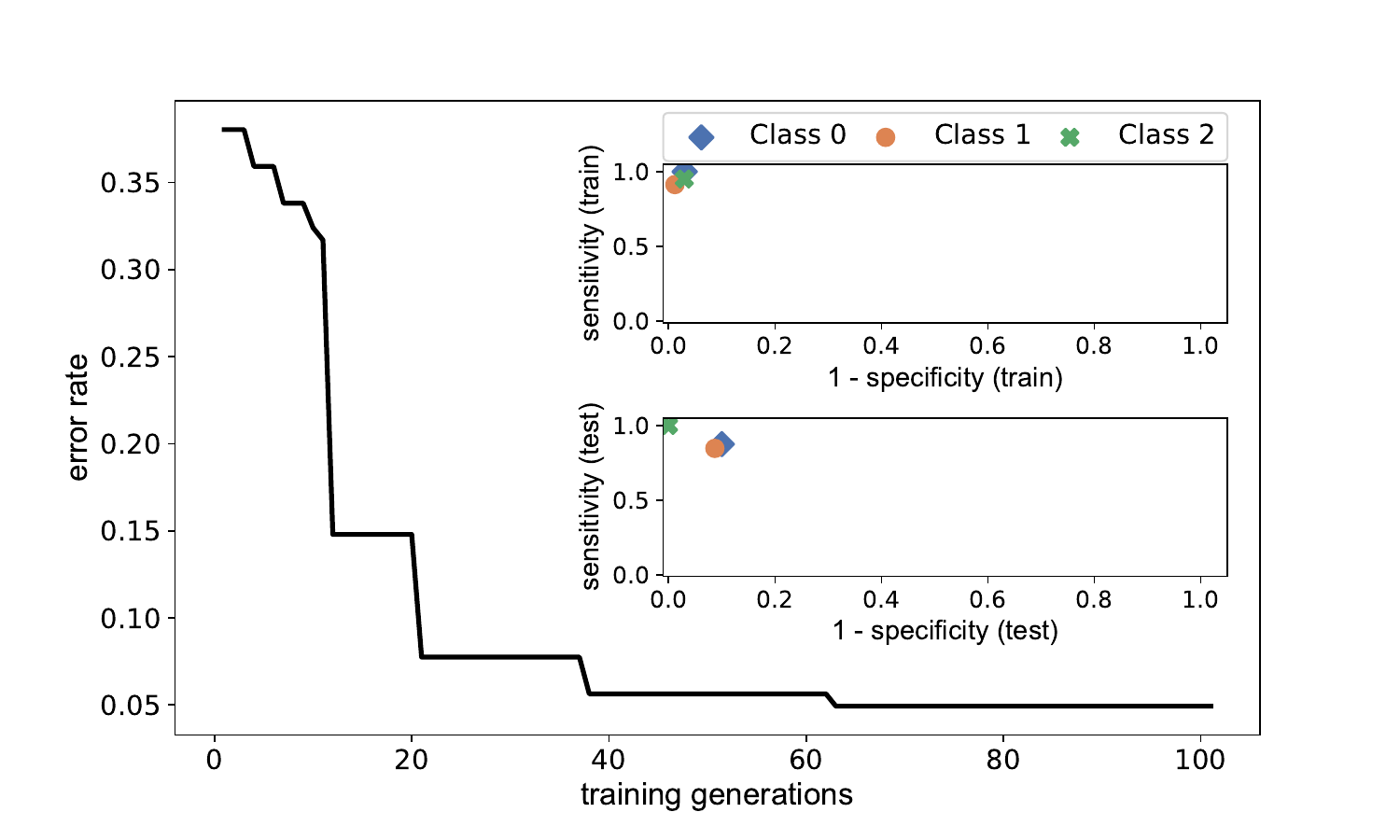}
    }
    
    \caption{Performance of an evolutionary generation for the Iris-T$_1$ (top) and Wine-T$_a$ (bottom) results shown in~Fig.~\ref{fig:nt_best_wine}. The error-rate reduction is shown by the line, the ROC plot for train (top) and test (bottom) in each plot.}
    \label{fig:nt_best_wine_itr}
\end{figure}

Fig.~\ref{fig:nt_best_wine} shows the best trees obtained by using hypervolume analysis for Iris (marked T$_1$ and T$_2$ in Fig.~\ref{fig:hv_plots}) and Wine (marked T$_a$ and T$_b$ in Fig.~\ref{fig:hv_plots}). Fig.~\ref{fig:nt_best_wine_itr} illustrate the example of MONT$_3$ learning ability over the 100 generations of the evolutionary optimisation process for datasets Iris and Wine. In Fig.~\ref{fig:nt_best_wine_itr}, the MONT$_3$ performance is also indicated through training and test receiver operating characteristic (ROC) curve plot where solutions lying top-left corner indicate good performance of the classifier.
%

Moreover, since hypervolume inductor analysis is the performance quantifier of the multi-objective algorithm's solutions quality, we compared MONT$_3$ solutions with MONT$_1$ and MONT$_2$ with a reference point bearing test error-rate 1.0 and tree-size 100. The results of the analysis are shown in Table~\ref{tab:hv_res}.
\begin{table}
    \begin{center}
        \renewcommand{\arraystretch}{1}
        \setlength{\tabcolsep}{10pt}
        {\caption{Quality of Trade-Off Obtained by Hypervolume Indicator $ H_i$ on Three Versions of MONT. }
            \label{tab:hv_res}}        
        \begin{tabular}{cccc}
            \hline
            Data & MONT$_1$ & MONT$_2$ & MONT$_3$\\
            \hline
            aus & 84.10 & 83.57 & 83.57\\
            gls & 44.14 & 56.16 & 52.35\\
            hrt & 78.74 & 77.46 & 77.46\\
            ion & 85.65 & 83.83 & 88.97\\
            irs & 85.60 & 90.00 & 89.97\\
            pma & 73.38 & 76.01 & 76.01\\
            vhl & 49.21 & 52.72 & 51.99\\
            win & 86.47 & 87.44 & 87.28\\
            wis & 90.68 & 90.55 & 91.34\\
            \hline
            Avg. & \textbf{75.33} & \textbf{77.53} & \textbf{77.66}\\
            \hline
        \end{tabular}
    \end{center}
\end{table}

\subsection{Neural Tree Performance Against Other Algorithms}
\label{sec:res_nt_vs_all}
The collected results of 30 runs of MONT$_3$ and 30 runs of the mentioned algorithms HFNT, MLP, REP-T, NBC, DT, and SVM are shown in Table~\ref{tab:nt_vs_all}. Table~\ref{tab:nt_vs_all} shows the average test error and variance over the mentioned nine datasets. Since not a single algorithm's performance (measured as per the average test error-rate) outperform all other algorithms, the three lowest average test error obtained by the respective algorithms for each dataset are marked in bold (Table~\ref{tab:nt_vs_all}). 

We performed two-sided \textit{t-test} statistics with setting the \textit{alpha} value to 0.05 to compare the average error-rate of MONT$_3$ algorithm against other algorithms for each dataset. Hence, as a null hypothesis, we test ``whether the average test error-rate of the MONT$_3$ is significantly lower than the average test error-rate of other algorithms?'' The results of the statistical t-test are shown in Table~\ref{tab:nt_vs_all_ttest}.  
\begin{table*}
    \begin{center}
        \renewcommand{\arraystretch}{1}
        \setlength{\tabcolsep}{10pt}
        {\caption{Average Test Error-Rate $ F_{\mu} $ and Variance $ F_{\sigma} $ of 30 Runs of Experiments on MONT$_3$ and Other Algorithms}
            \label{tab:nt_vs_all}}
        \begin{tabular}{lrrrrrrrrrrr}
            \hline
            & & \multicolumn{10}{c}{data} \\
            \cline{3-11}
            Algorithm &  \multicolumn{1}{c}{$ f_1 $} & 
            \multicolumn{1}{c}{aus}   & \multicolumn{1}{c}{hrt}  &  \multicolumn{1}{c}{ion}   & \multicolumn{1}{c}{pma}  &  \multicolumn{1}{c}{wis}   & \multicolumn{1}{c}{irs}  & \multicolumn{1}{c}{win}   & \multicolumn{1}{c}{vhl}  & \multicolumn{1}{c}{gls}   & \multicolumn{1}{c}{Avg.} \\
            \hline
            MONT$_3$ & $ f_{\mu} $ & \textbf{0.111} & \textbf{0.191} & \textbf{0.102} & \textbf{0.201} & \textbf{0.038} & \textbf{0.011} & \textbf{0.048} & \textbf{0.450} & 0.371 & \textbf{0.169} \\
            & $ f_{\sigma} $ & 0.002 & 0.000 & 0.000 & 0.000 & 0.000 & 0.000 & 0.000 & 0.003 & 0.001 & 0.021 \\
            HFNT & $ f_{\mu} $ & 0.174 & 0.230 & 0.178 & 0.284 & 0.065 & 0.189 & 0.176 & 0.591 & 0.601 & 0.276 \\
            & $ f_{\sigma} $ & 0.006 & 0.004 & 0.003 & 0.003 & 0.001 & 0.019 & 0.014 & 0.005 & 0.015 & 0.039 \\
            MLP & $ f_{\mu} $ & 0.175 & \textbf{0.213} & \textbf{0.094} & \textbf{0.249} & \textbf{0.024} & \textbf{0.040} & \textbf{0.037} & \textbf{0.183} & 0.367 & \textbf{0.154} \\
            & $ f_{\sigma} $ & 0.001 & 0.004 & 0.001 & 0.001 & 0.001 & 0.002 & 0.000 & 0.001 & 0.004 & 0.013 \\
            REP-T & $ f_{\mu} $ & \textbf{0.150} & 0.247 & 0.107 & 0.255 & 0.096 & 0.064 & 0.071 & \textbf{0.291} & \textbf{0.348} & \textbf{0.181} \\
            & $ f_{\sigma} $ & 0.001 & 0.004 & 0.002 & 0.001 & 0.003 & 0.001 & 0.000 & 0.001 & 0.005 & 0.012 \\
            NBC & $ f_{\mu} $ & 0.231 & \textbf{0.176} & 0.166 & \textbf{0.244} & \textbf{0.026} & 0.047 & \textbf{0.070} & 0.544 & 0.525 & 0.225 \\
            & $ f_{\sigma} $ & 0.001 & 0.003 & 0.002 & 0.001 & 0.001 & 0.001 & 0.001 & 0.001 & 0.008 & 0.035 \\
            DT & $ f_{\mu} $ & \textbf{0.146} & 0.312 & 0.126 & 0.337 & 0.514 & 0.070 & 0.370 & 0.463 & \textbf{0.337} & 0.297 \\  
            & $ f_{\sigma} $ & 0.001 & 0.012 & 0.001 & 0.001 & 0.004 & 0.001 & 0.002 & 0.002 & 0.006 & 0.024 \\
            SVM & $ f_{\mu} $ & 0.455 & 0.461 & \textbf{0.073} & 0.353 & 0.532 & \textbf{0.029} & 0.369 & 0.754 & \textbf{0.353} & 0.376 \\
            & $ f_{\sigma} $ & 0.002 & 0.004 & 0.001 & 0.001 & 0.006 & 0.001 & 0.002 & 0.000 & 0.004 & 0.046 \\
            \hline
            \\[-6pt]
            \multicolumn{12}{p{16cm}}{\textbf{Note:} for all datasets three lowest average test error rates are marked in Bold.} 
        \end{tabular}
    \end{center}
\end{table*}

Table~\ref{tab:nt_vs_all_best} shows the best test error of 30 runs of experiments over the mentioned nine datasets. Similar to the performance of algorithms shown in Table~\ref{tab:nt_vs_all}, the three lowest test error-rate by respective algorithms for each dataset are marked in bold (Table~\ref{tab:nt_vs_all_best}). Section~\ref{sec:dis} presents a detailed discussion of the performance of the algorithms.

\begin{table}
    \begin{center}
        \renewcommand{\arraystretch}{1}
        \setlength{\tabcolsep}{5pt}
        {\caption{Two-Sided T-Test: MONT$_3$ Against Other Algorithms}
            \label{tab:nt_vs_all_ttest}}
        \begin{tabular}{llrrrrrr}%
            \hline
            Data & T-test & HFNT & MLP & REP-T & NBC & DT & SVM \\
            \hline
            aus & t-stat & -7.02 & -4.59 & -13.52 & -19.20 & -6.42 & -44.81 \\
             & p-Value & 0.00 & 0.00 & 0.00 & 0.00 & 0.00 & 0.00 \\
            hrt & t-stat & -6.69 & -4.31 & -2.98 & \textbf{0.00}$*$ & -5.94 & -22.71 \\
             & p-Value & 0.00 & 0.00 & 0.00 & \textbf{1.00}$*$ & 0.00 & 0.00 \\
            ion & t-stat & -3.14 & -7.68 & \textbf{0.91}$*$ & -7.73 & -0.78 & 3.67 \\
             & p-Value & 0.00 & 0.00 & \textbf{0.36}$*$ & 0.00 & 0.44 & 0.00 \\
            pma & t-stat & -22.68 & -8.55 & -9.04 & -7.57 & -8.28 & -25.65 \\
             & p-Value & 0.00 & 0.00 & 0.00 & 0.00 & 0.00 & 0.00 \\
            wis & t-stat & 0.00 & 0.00 & 0.00 & 0.00 & 0.00 & 0.00 \\
             & p-Value & 0.00 & 0.00 & 0.00 & 0.00 & 0.00 & 0.00 \\
            irs & t-stat & -8.24 & -7.06 & -3.79 & -4.92 & -7.25 & -2.97 \\
            & p-Value & 0.00 & 0.00 & 0.00 & 0.00 & 0.00 & 0.00 \\
            win & t-stat & -39.64 & -5.98 & \textbf{3.91}$\dagger $ & \textbf{2.93}$\dagger $ & -4.85 & -33.49 \\
             & p-Value & 0.00 & 0.00 & \textbf{0.00}$\dagger $ & \textbf{0.00}$\dagger $ & 0.00 & 0.00 \\
            vhl & t-stat & -3.01 & -11.17 & 39.23 & -16.33 & 21.61 & -56.72 \\
             & p-Value & 0.00 & 0.00 & 0.00 & 0.00 & 0.22 & 0.00 \\
            gls & t-stat & \textbf{1.85}$*$ & -10.45 & -0.18 & -9.51 & \textbf{1.18}$*$ & \textbf{0.84}$*$ \\
             & p-Value & \textbf{0.07}$*$ & 0.00 & 0.86 & 0.00 & \textbf{0.24}$*$ & \textbf{0.40}$*$ \\ 
            \hline
            \\[-6pt]
            \multicolumn{8}{p{8cm}}{\textbf{Note:} For all datasets \textbf{NOT} marked in \textbf{Bold}, the MONT's average test error-rate was statistically significant than listed algorithms. For results marked $ * $, other algorithm's mean was better than MONT, but statistically \textbf{NOT} significant.  Only for dataset ``win'' indicated with symbol $ \dagger $, REP-Tree and NBC are statistically significant.}
        \end{tabular}
    \end{center}
\end{table}

\begin{table}
    \begin{center}
        \renewcommand{\arraystretch}{1.1}
        \setlength{\tabcolsep}{5pt}
        {\caption{Best Test Error-Rate of 30 Runs of Experiments on MON$_3$ Compared With Other Algorithms}
            \label{tab:nt_vs_all_best}}
        \begin{tabular}{lccccccc}
            \hline
            &    \multicolumn{7}{c}{algorithm} \\
            \cline{2-8}
            Data  & MONT$_3$    & HFNT & MLP  & REP-T      & NBC    & DT     & SVM   \\
            \hline
            aus    &  \textbf{0.101} & 0.072 & 0.123 &  \textbf{0.087}  & 0.181 &  \textbf{0.094}  & 0.326 \\
            hrt    &  \textbf{0.093}  & 0.111 & \textbf{0.074}  & 0.148 &  \textbf{0.074}  & 0.167 & 0.315 \\
            ion    & 0.042 & 0.070 & \textbf{0.028}  & 0.056 & 0.085 &  \textbf{0.028}  &  \textbf{0.028}   \\
            pma    &  \textbf{0.182} & 0.195 & 0.201 &  \textbf{0.195}  &  \textbf{0.182}  & 0.293 & 0.292 \\
            irs   & \textbf{0.000} & 0.000 & 0.000 & 0.000 & 0.000 & 0.000 & 0.000 \\
            wis  &  0.018 & \textbf{0.009} &  \textbf{0.000} & 0.028 &  \textbf{0.000} & 0.345 & 0.361 \\
            win    &  \textbf{0.000} & 0.000 &  \textbf{0.009}  &  \textbf{0.026}  & 0.027 & 0.254 & 0.281 \\
            vhl    &  \textbf{0.388} & 0.447 &  \textbf{0.147}  &  \textbf{0.218}  & 0.482 & 0.400 & 0.712 \\
            gls    & 0.302 & 0.372 & 0.256 &  \textbf{0.209}  & 0.395 &  \textbf{0.209}  &  \textbf{0.209}   \\	
            \hline
            \\[-6pt]
            \multicolumn{7}{l}{\textbf{Note:} the best three are marked in Bold}
        \end{tabular}
    \end{center}
\end{table}

\subsection{Neural Tree Training Versions Performances}
\label{sec:res_nt_vs_nt}
\begin{figure*}
    \centering
    \centerline{\includegraphics[scale=0.5]{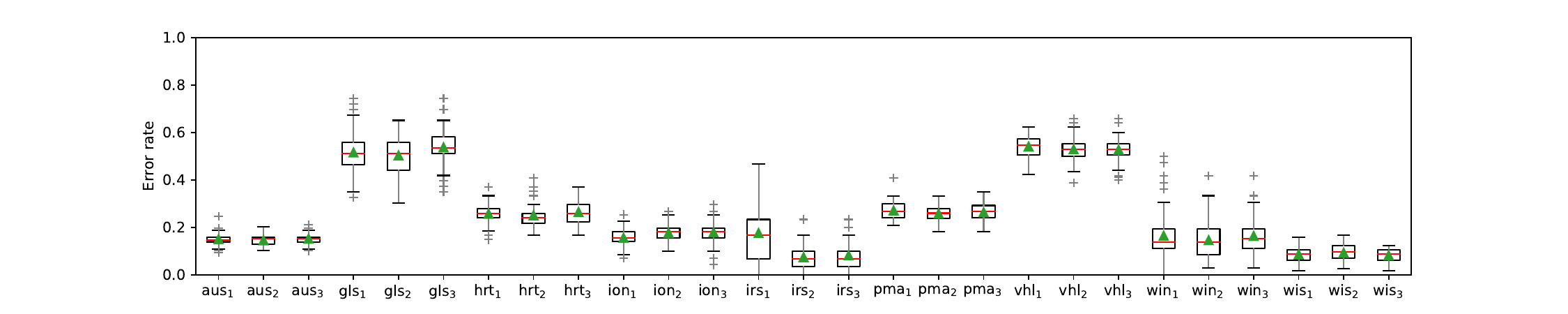}}
    (a)
    \centerline{\includegraphics[scale=0.5]{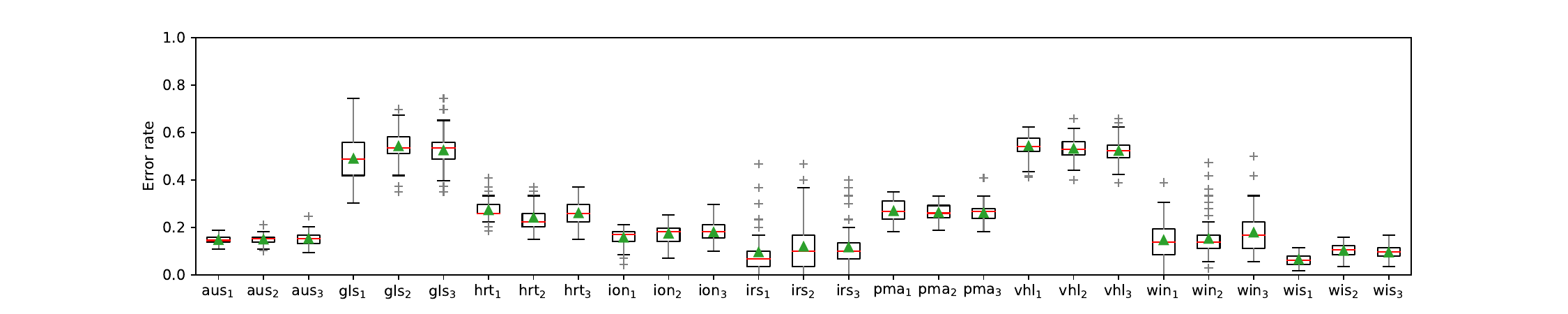}}
    (b)
    \caption{Performance of neural tree training versions: (a) Performance of optimisers GP, NSGA-II, and NSGA-III respectively marked 1, 2, and 3 as the subscript of the dataset names. (b) Performance of for activation functions at neural nodes: Gaussian, sigmoid, and tanh respectively marked 1, 2, and 3 as the subscript of dataset names. The median of error-rates is marked in red; the average error-rate is marked in a green triangle.} 
    \label{fig:nt_vs_nt_algo}
\end{figure*}

Apart from comparing the MONT$_3$'s performance against other algorithms, the performances of the MONT's training versions were also evaluated and compared. Fig.~\ref{fig:nt_vs_nt_algo} shows the performance of three training versions of MONT: MONT$_1$, MONT$_2$, and MONT$_3$. Fig.~\ref{fig:nt_vs_nt_algo}(a) is a box plot of the error-rates collected for 30 runs of each of these training versions, i.e. optimisation of MONT using GP, NSGA-II, and NSGA-III respectively. Additionally, Fig.~\ref{fig:nt_vs_nt_algo}(b) shows the performance of MONT for three versions of nodes used: Gaussian, sigmoid, and tangent hyperbolic labelled 1, 2, and 3, respectively.    

Moreover, the average training error-rates and average tree-size produced by  MONT$_1$, MONT$_2$, and MONT$_3$ representing training by GP, NSGA-II, and NSGA-III respectively are shown in Table~\ref{tab:tree_error_size}.

\begin{table}
    \begin{center}
        \renewcommand{\arraystretch}{1.1}
        \setlength{\tabcolsep}{1pt}
        {\caption{Average Training Error and Tree-Size Obtained by Three Versions of MONT Training and HFNT}
            \label{tab:tree_error_size}}
        \begin{tabular}{lccccccccc}
            \hline
            &  \multicolumn{4}{c}{Avg. training error-rate} & & \multicolumn{4}{c}{Avg. tree-size}\\
            \cline{2-5}  \cline{7-10} 
            data & MONT$_1$ & MONT$_2$ & MONT$_3$ & HFNT & & MONT$_1$ & MONT$_2$ & MONT$_3$  & HFNT \\
            \hline
            aus & 0.15 & 0.15 & 0.15 & 0.072 & &  20.03 & 8.43 & 8.13 & 8.53 \\
            hrt & 0.26 & 0.25 & 0.26 & 0.111 & & 27.20 & 8.59 & 8.73 & 9.20 \\
            ion & 0.16 & 0.18 & 0.18 & 0.070 & & 32.16 & 9.6 & 9.83 & 7.63 \\
            pma & 0.27 & 0.26 & 0.26 & 0.195 & & 49.28 & 10.61 & 10.28 & 14.00 \\
            irs & 0.18 & 0.07 & 0.08 & 0.000 & & 218.51 & 17.07 & 16.13 & 156.73 \\
            wis & 0.09 & 0.09 & 0.08 & 0.009 & & 30.49 & 9.2 & 9.91 & 8.97 \\
            win & 0.16 & 0.15 & 0.16 & 0.000 & & 44.37 & 13.78 & 13.94 & 26.20 \\
            vhl & 0.54 & 0.53 & 0.53 & 0.447 & & 51.36 & 17.74 & 18.13 & 31.23 \\
            gls & 0.52 & 0.5 & 0.54 & 0.372 & & 85.53 & 25.87 & 25.83 & 60.20 \\
            \hline
            Avg. & \textbf{0.26} & \textbf{0.24} & \textbf{0.25} & \textbf{0.276} & & \textbf{62.37} & \textbf{13.43} & \textbf{13.42} & \textbf{35.85} \\
            \hline
        \end{tabular}
    \end{center}
\end{table}

\section{Discussion}
\label{sec:dis}
The MONT optimisation is an evolutionary process where a population of MONTs competes to yield the fittest solution. The computational properties of a MONT are similar to an MLP except for that the MONT has a tree-like structure compared to MLP that has fully connected network-like structure and MONT does a simultaneous feature selection ignoring insignificant features during its induction.

The MONT uses an evolutionary learning process for its tree structure evolution. The optimisation of tree-structure gives MONT the ability to find a hypothesis from a large hypothesis search-space as mentioned in Section~\ref{sec:mo_ec}. Moreover, two multi-objective optimisers NSGA-II and NSGA-III both bearing differing mechanism for maintaining diversity in the population of an evolutionary process ensures exploration of this large hypothesis search space. This gives MONT the ability to induce a hypothesis from topological-space, feature-space, and parameter-space specific to each class of each problem.

Table~\ref{tab:nt_vs_all} shows that the performance of the solution obtained by hypervolume inductor analysis from MONT$_3$'s population dominates other listed algorithms, which is evident from the MONT$_3$ being in the top three performing algorithms for the most selected datasets. The average test error-rate of the MONT$_3$ outperformed all algorithms for three datasets: Australia, Iris, and Pima (Table~\ref{tab:nt_vs_all}). The performance of MONT$_3$ was closely second to other algorithms for the datasets Wine, Wisconsin, and Vehicle. For dataset Glass, the performance of all algorithms marginally differed from each other, and the best performing algorithms was REP-T. 

The statistical significance test using two-sided t-test shown in Table~\ref{tab:nt_vs_all_ttest} indicates that MONT$_3$ has statistically significant performance for dataset Australia, Iris, and Pima, and Wisconsin. For datasets, Heart, Ionosphere, Vehicle, can be said competitive since some other algorithm's average test error rates were better but not statistically significant. For the dataset, wine, however, REP-Tree and NBC appeared to be statistically significant.  

Similarly, the best test error-rates of the algorithms over datasets shown in Table~\ref{tab:nt_vs_all_best} indicate that MONT$_3$ performed well on six datasets out of nine datasets. And, MONT$_3$ performed competitively over two datasets. This is evident from the statistical test where the average error-rate of MONT was found statistically significant and equivalent to 8 datasets. This performance of the MONT$_3$ is in the view that MONT$_3$ induce a hypothesis by simultaneous minimisation of the hypostasis complexity (model's parameter reduction)  and the error rate minimisation.  Whereas, other algorithms had a single objective (error rate) to minimise. However, one advantage with MONT$_3$ its nature of being population-based hypothesis induction compared with other mentioned algorithm. Consider this fact the MONT$_3$ was trained with small-scale training set with a minimum of 100 iterations and a population of 50 individuals. Hence, the performance of MONT$_3$ may improve for a higher number of iteration and population size.

Apart from comparing MONT$_3$ (trained by NSGA-III) performance against other algorithms, its performance was compared  with two other training versions: MONT$_2$ (trained by NSGA-II) and single objective (optimisation of error-rates) version of MONT trained with GP.  As per the obtained results of these three training versions shown in Fig.~\ref{fig:nt_vs_nt_algo}(a), the NSGA-III based optimisation of MONT, i.e., version MONT$_3$ performed well in the cases of six datasets. As well as, the Pareto-optimality analysis shown in Table~\ref{tab:hv_res} shows that on a hypervolume indicator analysis, the Pareto-optimal solutions set generated by NSGA-III is competitively better than NSGA-II and GP, which is attributed to its exploitation of non-dominated sorting and niching operator for population diversity maintenance. This is evident from average $ H_i $ values 77.66 obtained for NSGA-III is higher than the NSGA-II ($ H_i = 77.53$) and GP ($ H_i = 75.33$). That is a solution in the NSGA-III tends to offer a better trade-off between tree-size and test error-rates. 

The MONT$_3$ performance was also compared for  three different types of activation functions: Gaussian, Sigmoid, and tangent hyperbolic that can be used as activation nodes. The performance of Gaussian function was found better in the cases of seven datasets [Fig.~\ref{fig:nt_vs_nt_algo}(b)]. However, the performance of the Sigmoid function was competitive and very close to the Gaussian function. The Gaussian function may have advantages of its ability to possess varies shapes during training, whereas the sigmoid function the advantages of the bias at the neural nodes.  Moreover, this is evident from MONT's better performance compared to HFNT that uses a set of heterogeneous nodes (a variety of activation function set) without a bias input (Table~\ref{tab:tree_error_size}). As well as, MONT$_3$'s NSGA-III based training results lower tree-size than HFNT.

In MONT algorithm, each class is represented by a subtree, and each subtree competes with another subtree to maximise its classification accuracy  (Fig.~\ref{fig:nt_best_wine_itr}). This comparative nature of subtrees of MONT results in maximisation of the MONT's  classification accuracy. This is evident from the differences between the subtrees of each class obtained by the MONT algorithm, as shown in Fig.~\ref{fig:nt_best_wine} for Iris and Wine. Moreover, the learning process of each class during an evolutionary process shown in Fig.~\ref{fig:nt_best_wine_itr}.

\section{Conclusion}
\label{sec:con}
This paper proposes a new algorithm so-called multi-output neural tree (MONT) that is trained by using non-dominated sorting algorithm frameworks (NSGA) III. The evolutionary process yields a population of MONTs as the solutions to a problem. The population of MONTs was analysed using hypervolume indicator analysis for selection of a Pareto-optimal set. The performance of the MONT was compared with five well-known algorithms: decision tree, multilayer perceptron, reduced error-pruning tree, na{\"{i}}ve Bayes, and support vector machine and heterogeneous flexible neural tree. The performance of MONT outperforms the mentioned algorithms on three datasets, and for another three datasets, it was competitive with other algorithms. In general, results show that the MONT performed competitively over a larger set of chosen data compared with the other algorithms. This performance of MONT attributed to its property where each class is represented as a subtree that competes against the other subtrees for improving the classification accuracy. In an additional set of experiments MONT's three training version GP, NSGA-II, and NSGA-III were compared, where NSGA-III emerged better in terms of minimising the trade-off between two objective tree-size and training accuracy. The use of activation function Guassun and sigmoid were found better choice compared to tangent-hyperbolic. Thus, simultaneously maximising overall classification accuracy and minimising tree-size (reducing parameter). These properties of MONT is crucial to achieving generalisation ability in searching a hypothesis from hypothesis-space.

\bibliographystyle{IEEEtran}
\bibliography{cec_wcci20}

\end{document}